\documentclass{article}

\usepackage[utf8]{inputenc}
\usepackage{booktabs} 
\usepackage{siunitx}  
\usepackage{bussproofs}
\usepackage{url}
\usepackage{graphicx}
\usepackage{amsmath}

\newcommand{\subs}{\sqsubseteq}  

\begin{document}


\title{Passing the Brazilian OAB Exam: data preparation and some experiments\thanks{The authors would like to thank João Alberto de Oliveira Lima for introducing us to the LexML resources, and Peter Bryant for his careful review of the article.}}



\author{Pedro Delfino\thanks{FGV Direito Rio and FGV/EMAp}
  \and Bruno Cuconato\thanks{FGV/EMAp}
  \and Edward Hermann Haeusler\thanks{PUC-Rio}
  \and Alexandre Rademaker\thanks{IBM Research and FGV/EMAp}}

\maketitle

\begin{abstract}
In Brazil, all legal professionals must demonstrate their knowledge of the law and its application by passing the OAB exams, the national bar exams. The OAB exams therefore provide an excellent benchmark for the performance of legal information systems since passing the exam would arguably signal that the system has acquired capacity of legal reasoning comparable to that of a human lawyer. This article describes the construction of a new data set and some preliminary experiments on it, treating the problem of finding the justification for the answers to questions. The results provide a baseline performance measure against which to evaluate future improvements. We discuss the reasons to the poor performance and propose next steps.
\end{abstract}



\section{Introduction}\label{sec:intro}

The ``Ordem dos Advogados do Brasil'' (OAB) is the professional body of lawyers in Brazil. Among other responsibilities, the institution is responsible for the regulation of the legal profession in the Brazilian jurisdiction. One of the key ways of regulating the legal practice is through the ``Exame Unificado da OAB'' (unified Bar examination).
Only those who have been approved on this exam are allowed to work as practising attorneys in the country. In this way, the OAB exam is similar to the US Bar Exam.

Thus, the OAB exam provides an excellent benchmark for the performance of a system attempting to reason about the law. That is, passing the OAB exam would signal that the system has acquired the most important aspects of legal knowledge, up to a level comparable to a human lawyer. 

This paper reports our effort in the construction of the data set and some preliminary experiments with the collected data. We obtained the official data from previous exams and their answer keys from \url{http://oab.fgv.br/}. 
As our first contribution, we collected the PDF files, extracted the text from them, and cleaned up the results, producing machine-readable data.~\footnote{All data files are freely available at \url{http://github.com/own-pt/oab-exams}.} In Section~\ref{sec:exam} we describe the data.


Question answering is an automatic way of determining the right answer to a question presented in natural language form \cite{mitkov2005oxford}. 
An ideal legal question answering system would take a question in natural language and a corpus of all legal documents in a given jurisdiction, and would return both a correct answer and its legal foundation, i.e., which sections of which norms provide support for the answer. Since such a system is far from our current capabilities, we started with a simpler task.

In \cite{Fawei:2016tg} the authors report a textual entailment study on this US Bar exam material. In the experiment, the authors treat the relationship between the question and the multiple-choice answers as a form of textual entailment. Answering a multiple choice legal exam is a more feasible challenge, although it is still a daunting project without restrictions on the input form, such as preprocessing natural language questions to make them more intelligible to the computer (see Section~\ref{sec:exam}) or restricting the legal domain. That is the reason we have chosen to restrict the domain to a single section of the OAB exams: Ethics, one which is governed by only a few legal norms. 

We have conducted three experiments in question answering (Section~\ref{sec:setup}). In the first experiment, we tried to find the right answer between the multiple-choice alternatives. The last two were in shallow question answering (SQA), a form of question answering where a system does not provide answers directly, but rather retrieves documents that justify the already provided answer. We have adapted the methodology described in \cite{monroy2008graphs,monroy2009nlp} to answer multiple-choice exams instead of closed-ended answers.


To be able to provide the right justification (the article of a law) for each question, we needed to have the text of the laws available. This is a particular challenge in the
legal domain, as normative instruments are not readily available in a uniform format, suitable for being consumed by a computer program. Fortunately, using resources provided by the LexML Brasil project, we were able to collect and convert to XML format all the normative documents we needed, which were initially provided in HTML and PDF (Section~\ref{sec:lexml}).

\section{The OAB Exams}\label{sec:exam}


Before 2010, Brazilian bar exams were regional -- each state in the country had its own exam. Only in 2010 were the exams nationally unified. 
In order to be approved, candidates need to be approved in two stages. The first phase consists of multiple choice questions.
The second phase involves free-text questions.
The first phase has 80 multiple choice questions and each question has 4 options, ranging from letters A to D. Candidates are asked to choose the correct alternative and in order to be approved, candidates need at least a 50\% performance.

Every year, there are 3 applications of the exam in the country, which are currently being conducted by FGV~\footnote{\url{http://fgvprojetos.fgv.br/node/135}}. Concerning exam statistics, the first phase is responsible for eliminating the majority of the candidates. Historically, the exam has a global 80\% failure rate. The most recent exam, whose multiple choice stage was held on July 2017, had the highest failure rate: 86\% of the candidates failed.

Since 2012, the exams have revealed a pattern for which areas of Law the examination board focuses on and in which order the questions appear on the exam.
Traditionally, the first 10 questions are about Ethics. Despite the name, this area is not about classic texts on morals from Aristotle or Socrates. In the context of the exam Ethics means questions about the rights, the duties and the responsibilities of the lawyer. This is the simplest part of the exam with respect to the legal foundation of the questions. Almost all the questions on Ethics are based on the Federal law 8906 from 1994, which is a relatively short (89 articles) and well designed normative document. A minor part of the questions on Ethics is related to two other norms: (i) ``Regulamento Geral da OAB'' (OAB General Regulation, 169 articles) and (ii) ``Código de Ética da OAB'' (OAB ethics code, 66 articles).
It is important to note that these two norms are neither legislative nor executive norms. Indeed, they are norms created by OAB itself. OAB's prerogative to do so is assured by 8906 federal law.

Beyond the first 10 questions about Ethics, the remaining questions appear in the order of areas presented in Table~\ref{tab:areas}, starting with Ethics and ending with taxes. The committee responsible for the exam may change this pattern at any time, since, in this aspect, they are not bound by the edict of the exam. 
\begin{table}[htbp]
\centering
\begin{tabular}{rcc|rcc}
\hline 
\textbf{area}  & \textbf{\#} & \textbf{(\%)} 
& \textbf{area}  & \textbf{\#} & \textbf{(\%)}\\\hline
Ethics  & 10 & 65 & Constitutional Law  & 7  & 42 \\
Consumer's Law &  2 & 56 & Civil Procedures &  6 & 40 \\
Children's Law &  2 & 54  & Philosophy &  2 & 40 \\
Criminal Procedures &  5 & 47 & Labor's Law Proc. &  6 & 40  \\
Regulatory Law &  6 & 47 & Criminal Law &  6  & 38 \\
Human Rights  & 3 & 47 & International Law & 2 & 37 \\
Civil Law &  7 & 44 & Business Law &  5 & 33 \\
Environmental &  2 & 43 & Taxes &  4 & 42 \\
Labor's Law & 5 & 42 & & & \\
\hline
\end{tabular}
\caption{Questions per subject area and their performance rates}\label{tab:areas}
\end{table}

In addition to the high frequency rate, the Ethics questions also have a high performance rate among candidates, while Business Law has the lowest performance rate. In the third column of Table~\ref{tab:areas}, we reproduce some descriptive statistics from \cite{oab-numeros-2016}. This column shows the average performance on the different areas covered by the exam. 

We obtained the previous exam files from the FGV website~\footnote{\url{http://oab.fgv.br/}.}. After downloading exam files, we converted them from PDF to text using Apache Tika~\footnote{\url{https://tika.apache.org/}.}. Finally, we manually revised the obtained text, and introduced markup to signal the beginning of questions, alternatives and some meta data about them. We have chosen a simple set of markup
to facilitate the final parser that converts the exam files to XML. This data cleanup was not easy, as it was necessary to deal with errors in the encoding of some files, missing words, hyphenated words and to remove irrelevant parts like footnotes, the instructions page etc. In order to make this research reproducible, all the data is available at our public repository on GitHub.~\footnote{\url{http://www.github.com/own-pt/oab-exams/}.}


The final data comprises 22 exams totaling 1820 questions. A range of issues on the texts of the questions of the exams was identified. Many of the problems are similar to the ones found in the BAR exams and described by \cite{Fawei:2016tg}. For instance, some questions do not contain an introductory paragraph defining a context situation for the question. Instead of that, they have only meta comments (e.g. ``assume that...'' and ``which of the following alternative is correct?'') followed by the choices. Some questions are in a negative form, asking the examinee to select the wrong option or providing a statement in the negative form such as ``The collective security order {\bf cannot} be filed by...''. 
%
%
%
%
Moreover, some questions explicitly mention the law under consideration, others do not. Many questions present a sentence fragment and ask for the best complement among the alternatives, also exposed as incomplete sentences. Nevertheless, we identify some similarities between questions of the same subject area. 

We sampled 30 questions on Ethics for analysis (from the 210 questions in all exams) and one of the authors manually identified the articles in the laws that justify the answer, creating our golden data set. The key finding was that, usually, one article on the federal law 8906 was enough to justify the answer to the questions (15 questions). Less often, the justification was not in the law 8906, but rather in OAB
General Regulation (3 questions), or on the OAB Ethics Code (8 questions). Three questions were justified by two articles in law 8906, and another in jurisprudence from the Superior Court of Justice about an article from the law 8906. 



\section{The LexML Brasil project}\label{sec:lexml}

For the experiments reported in Section~\ref{sec:results} and the further experiments that we plan to conduct, we needed three norms in a machine readable format: Federal law 8906 from 1994, the ``Regulamento Geral da OAB'' (OAB general regulation)  and the ``Código de Ética da OAB'' (OAB ethics code). Moreover, we needed the documents in a format that preserved the original internal structure of the documents, i.e., the sections, articles, and paragraphs.

The LexML~\cite{lexml} is a joint initiative of the Civil Law legal system countries seeking to establish open standards for the interchange, identification and structuring of legislative and court information. 
One of the initial goals of the initiative, later abandoned, was the standardization of a single document format (called LexML) for encoding normative documents of all participating countries. Currently only Brazil has developed a XML schema called ``LexML Brasil''. The remain participants migrated to Akoma Ntoso \cite{akoma2015} and EUR-Lex~\footnote{\url{http://eur-lex.europa.eu}.}. 

The LexML project has a public repository at \url{https://github.com/lexml}. As the repository shows, the project is active but it does not have vast documentation. One useful tool developed by the LexML Brazil team is a parser of legal documents~\footnote{\url{https://github.com/lexml/lexml-parser-projeto-lei}}, still in beta. The software receives as input a DOCX file (a common format for Brazilian legal documents) with the norm and outputs it in XML format, using the tags and the structure according to the conventions of the LexML schema. We had to make minor modifications in the three documents before submitting them to the parser; the XML files produced are available in our repository.


\section{The Experiments Setup}\label{sec:setup}

The original idea for the experiment is described in \cite{monroy2008graphs}, and it runs as follows: one collects the legal norms in a corpus and preprocesses them performing tasks such as converting text to lower case, eliminating punctuation and numbers and, optionally, removing stop-words. After that, the articles of the norms are represented as TF-IDF vectors in a Vector Space Model (VSM)~\cite{christopher2008introduction}.

In a VSM, every document $d$ is represented by a vector whose size is the vocabulary size of the corpus $D$. When using TF-IDF weighting, the value of each component $t$ of the vector corresponding to $d$ is given by Equation~\ref{eq:tfidf}.

\begin{equation}\label{eq:tfidf}
\mathrm{TFIDF}_{t, d} = \frac{f_{t, d}}{\sum_{t' \in d} f_{t',d}} 
 \log\left(\frac{|D|}{|d \in D: t \in d|}\right)\\
\end{equation}
where $f_{t,d}$ is the number of times term $t$ appears in document $d$.

A directed graph is then created, with a node for each article of a norm in the corpus. 
When provided a question-answer pair, our system preprocesses the question statement and the alternatives in the same way as it does to the articles in the base graph. It turns them into TF-IDF vectors using IDF values from the document corpus.\footnote{This means that if a term occurs in the question statement or alternative but not on the legal norm corpus, its IDF value is 0.} The statement node is connected to every article node, and each article node is then connected to every alternative node. In this we differ from \cite{monroy2008graphs}, as we have no need for heuristic rules for splitting the question provided.

The edges are given weights whose value is the inverse cosine similarity between the connected nodes' TF-IDF vectors. The system then calculates the shortest path between question statement and answer item using Dijkstra's algorithm, and returns the article that connects them as the answer justification. Unlike \cite{monroy2008graphs} our graph structure does not allow for more than one node connecting statement and alternative, as we knew from previous legal analysis that questions were usually justified by a single article.

The intuition behind such a method is that the more similar two nodes are, the lesser is the distance between them; as a document that answers a given query is presupposed similar to the question, it makes sense to retrieve the article in the shortest path between the statement and the alternative as a justification for the answer.

\begin{figure}
    \centering
    \includegraphics[width=.85\textwidth]{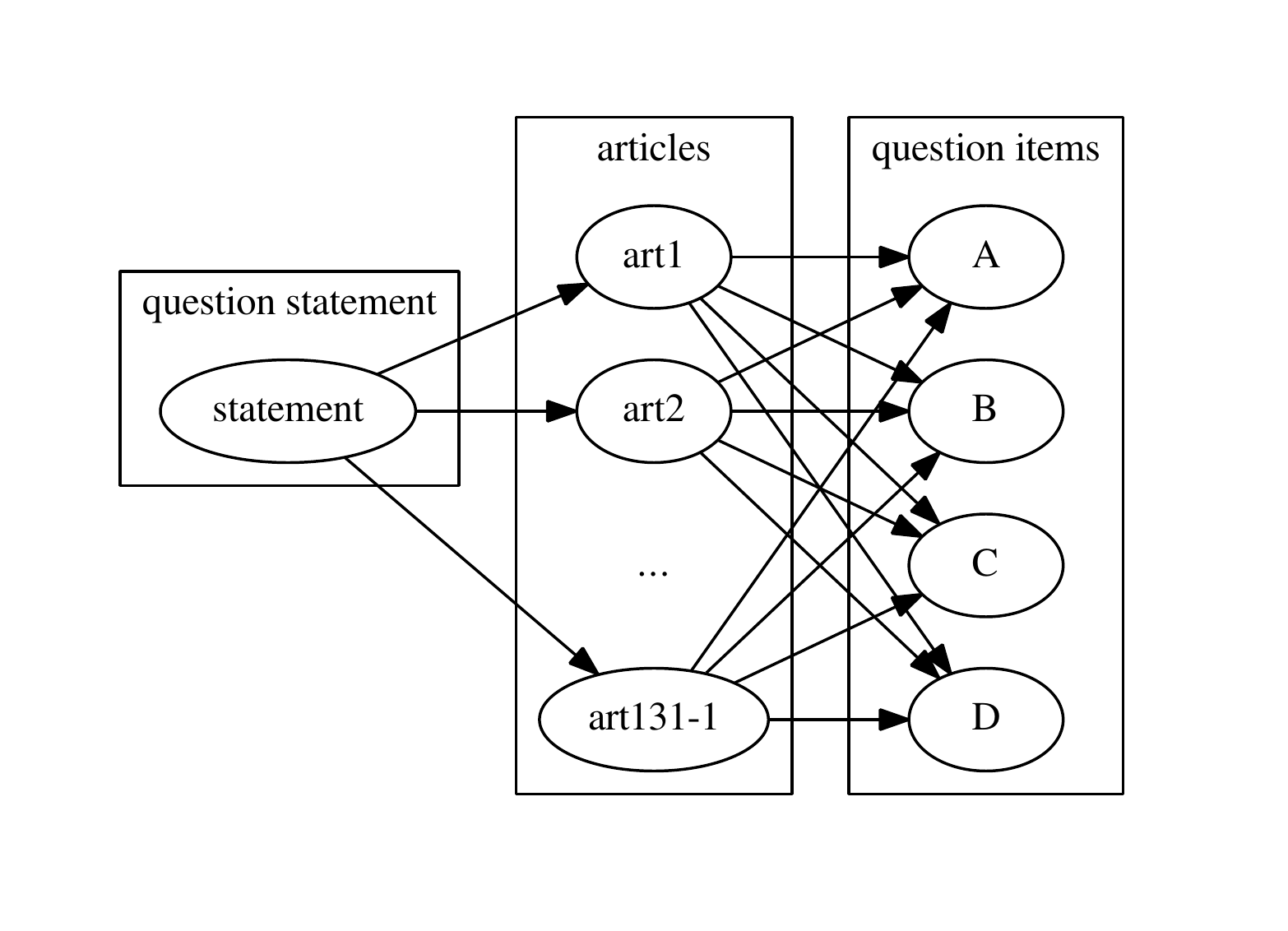}
    \caption{Graph layout of the SQA system. If a $A$ is the number of article nodes, we then have $5A$ edges, as we have one statement and four alternatives.}
    \label{fig:graph}
\end{figure}

\section{The Results}\label{sec:results}

We have taken the method described in Section~\ref{sec:setup} as a baseline for further improvements, and we have conducted three experiments using it. Our golden answer set was manually created by one of the authors (Section~\ref{sec:exam}). It consists of a table that includes 30 questions from eleven different editions of the OAB exam. Each line of the table represents one question, with fields for the year/edition of the exam where it appears, the question number, the URN (uniform resource locator) of the legal norm and its article(s) which support the answer.\footnote{This document is also available from our public repository.}

In our first experiment we had an ambitious objective: we had our system receive a question statement and its multiple alternatives, and we wanted it to retrieve the right answer along with its justification in the legal norm. When given the question and its alternatives, the system would add them to the base graph composed by the respective legal norm's articles, connecting them to the article nodes and the choices, but not to each other. The system would return 
the shortest paths 
between the question statement and its alternatives, and the presumed justification would be the article connecting the statement and the closest item to it. The system's performance against this task was not impressive: although it chose the correct alternative 10 times, 
it only provided the correct justification for 8 of these.

Analyzing the system's output paints a more nuanced picture, however. In some cases, the system would find the correct justification article for the correct answer, but would pick as its putative answer another (incorrect) item, because it had a shorter path. Other times, it would not be capable of deciding between two (or more) answer items, as they all had a shortest path of the same distance. The following exam question is a sample case where this statistical approach to question answering is defective:

\begin{quote}
    The young adults Rodrigo (30-year-old), and Bibiana (35-year-old), who are properly enrolled in an OAB section [\ldots]
    Considering the situation described, choose the correct alternative:

    A) Only {\bf Bibiana} meets the eligibility criteria for the roles.

    B) Only {\bf Rodrigo} meets the eligibility criteria for the roles.
    
    [\ldots]
    
    \hfill (2016 OAB exam, 19th edition, question 7)
\end{quote}

As one can see, these two options differ by only one word (the names of the fictional lawyers), and both  are unlikely to be in the text of the legal norms, which means that they do not affect the calculation of similarity. This gives us the same distance between the question statement and these two answer choices, and the system is incapable of choosing between them. A similar situation arises when one answer item makes a statement and another item denies this statement:

\begin{quote}
    [question statement]

    A) does {\bf not compel} him to pay the agreed upon legal fees.
    
    B) does {\bf compel} him to pay the agreed upon legal fees.
    
    [\ldots]
    
    \hfill (2015 OAB exam, 18th edition, question 1)
\end{quote}

In a question like this a system can only systematically report a correct answer if it has a higher-level understanding of the texts at hand: no bag-of-words model will suffice. In Section~\ref{sec:logic} we begin exploring logical approaches to question answering, which may offer a solution to this problem.

As our first experiment demonstrated that our simple system could not reliably pick the correct answer among four alternatives, we turned our attention from pure question answering to shallow question answering, where our system would only have to provide the correct legal basis for the correct answer provided along with the question. 

In our second experiment we read the three XML files corresponding to the norms in the Legal Ethics domain, and built separate base graphs from their articles. For each question in our golden set, we added its statement and its correct answer to the base graph built from the norm which justifies the answer.
The sole task of our system in this case is to the determine which article from the norm it was given that justifies the answer. In this simpler form, performance was not bad: the system retrieved the correct article in 21 out of 30 question-answer pairs.

In our third experiment, we tried to see if our system could provide the correct article (from the appropriate legal norm) without us telling it at which norm it should look. Following this idea, we have taken the articles from all norms and built a single base graph. For each question in our golden set, we again added its statement and correct answer as nodes connected to all article nodes in the graph (see Figure~\ref{fig:graph})
, and then calculated the shortest path between them to retrieve the system's putative justification to the question-answer pair. The system now had to retrieve the correct article among articles from all norms -- which, being in the same legal domain, had similar wordings and topics -- therefore increasing the difficulty of the task. Despite this, its performance did not plunge: it scored the right article in 18 out of the 30 question-answer pairs.

Although the results from the SQA experiments were not bad, they are not very encouraging. We would like to have an autonomous system that could pass the multiple-choice OAB exam, not an assistant system that could show us where the answer keys are legally grounded.

\section{A possible logic-based approach} \label{sec:logic}

One of the key observations that emerge from the results in Section~\ref{sec:results} is the importance of logical reasoning for our final goal of constructing a system to pass the OAB exam with a full understanding of the questions and laws. 

For the future, we aim to investigate how to enrich the data with lexical information and syntactic dependencies as an intermediary step toward a semantic representation of the questions and laws statements. Nevertheless, we have to decide what should be an adequate logic language to represent laws and the deep 
semantics from the text statements. Since the adequacy of a logic language can be evaluated even before a procedure to obtain logic expressions from natural language texts is developed, this section presents some preliminary discussion about one possible logic, the $i\mathcal{ALC}$ logic, which is a logic specifically developed to formalize laws and to reason regarding them \cite{HPR:2010a}. Here we sketch the ideas of how $i\mathcal{ALC}$ can be used in the further experiments with the OAB exams.

In \cite{HPR:2010a,HPR:2010b} we discuss how Kelsen's \cite{Kelsen1991} pure theory of law points out a framework that takes into account the legal knowledge forming a collection of individual, legally valid statements. Thus, each legally valid statement may be seen as an inhabitant among the many individual laws of the represented legal system. The natural precedence existing between individual legal statements can be taken as a pre-order relation on the legal statements. For example, ``Ana is liable'' precedes ``Ana signed a selling contract'', since in order to be valid the parts involved in the contract must be liable.

The legal principle that rules the stability of the law implies that the precedence of individual laws preserves properties (decisions, conditions of applicability, adequate fora, etc) regarding them. From these jurisprudence considerations we have that intuitionistic logic is the best choice for reasoning on laws within a non-deontic approach. In \cite{HPR:2010a,HPR:2010b} it is shown, by means of an example, how the intuitionistic negation is better than its classic counterpart for reasoning with conflict of laws.

In the presence of this natural precedence order between legally valid statements, the intuitionistic interpretation of subsumption between concepts $A$ and $B$ ($A\subs B$) reflects more adequately the structure of existing legal systems than its classical interpretation counterpart. The classical interpretation of $A\subs B$ says that any legally valid statement satisfying concept $A$ also satisfies concept $B$, no matter how these concepts are related to the precedence relation. For example, in the three tiers involved in the U.S. legal system (District court, Appellate court and Supreme court) the legally valid statements are strongly related by natural precedence order. If $E$ is taken to be a concept meaning {\em Environmental} crimes and $G$ means {\em Grave} crimes, then $E\subseteq G$ may not hold at all. A district court would not judge a situation of an oil-company that spoiled a beach in town as a grave crime. Note that at the federal level, the spoiling of natural resources is a grave crime. When considering the intuitionistic interpretation the subsumption $E\subseteq G$ holds, for any valid legal statement superseding ``oil-company spoiled the coast and is sentenced to provide a compensation to the county'' is superseded by a federal, valid legal statement classified as a grave crime. In the Supreme court, the oil-company would be sentenced for committing a grave crime.

$\mathcal{ALC}$~\cite{Baa03} is the basic description logic $\mathcal{AL}$ (for Attributive Language) plus the notion of complement $\mathcal{C}$ (negation) of arbitrary concepts. $\mathcal{ALC}$ is a fragment of classical first-order logic having only unary and binary relation symbols. $i\mathcal{ALC}$ is closed related to many known intuitionistic modal and hybrid logics, as is discussed in \cite{ialc-arxiv} and in \cite{PHR:2010}.

To illustrate the use of $i\mathcal{ALC}$ for reasoning over the OAB exams questions, let us consider the translated question below:

\begin{quote}
  Three friends graduated in a Law School in the same class: Luana, Leonardo, and Bruno. Luana, 35 years old, was already a manager in a bank when she graduated. Leonardo, 30 years, is mayor of the municipality of Pontal. Bruno, 28 years old, is a military policeman in the same municipality. The three want to practice law in the private sector. Considering the incompatibilities and impediments to practice, please select the correct answer.\\
 
  A) Luana is not prohibited from practicing law because she is an employee of a private institution, so there are no impediments or incompatibilities.

  B) Bruno, like all other civil servants, is only prohibited from practicing the law against the government agency that remunerates him.

  C) The three graduates, Luana, Leonardo, and Bruno, have
  functions incompatible with legal practice. They are therefore prohibited from exercising private practice. (CORRECT)

  D) Leonardo is banned from practicing law only against or in favor of legal entities from the public sector, public companies, mixed-capital companies, public foundations, parastatal entities or concessionaire corporations or public service licensees.
  
  \hfill (2016 OAB exam, 19th edition, question 4)
\end{quote}

The justification of the answer to this question is obtained in the law 8906, article 28.~\footnote{The complete text of the law can be found at \url{http://bit.ly/29gZc83}} The relevant fragments of this article, translated into English, are:

\begin{quote}
  Legal practice is incompatible, even for self-defense, with the following activities:

  I - head of the Executive and members of the Bureau of the
  Legislative Branch and their legal substitutes; [\ldots]

  V - occupants of positions or functions linked directly or
  indirectly the police activity of any nature; [\ldots]

  VIII - occupiers of management positions in financial institutions, including private ones. [\ldots]
\end{quote}

In $i\mathcal{ALC}$, the law 8906 is formalized as a concept defined as the intersection of the concepts from its articles, that is, $Law_{8906} \equiv Art_1 \sqcap\ldots\sqcap Art_{28} \sqcap\ldots\sqcap Art_{87}$. Article 28 in turn is also further formalized as the intersection of the concepts from its paragraphs, $Art_{28} \equiv P_1 \sqcap P_2 \ldots$. The paragraph VIII is formalized by the two concepts $Lawyer \subs\neg Financial$ and $Financial\subs\neg Lawyer$. Paragraph V is formalized by $Lawyer\subs\neg Police$ and $Police\subs Lawyer$. Finally, paragraph I by $Lawyer\subs\neg ChiefCouncil$ and $ChiefCouncil\subs\neg Lawyer$. The $Lawyer$ concept can be read as the set of valid legal statements (VLS) about lawyers. That is, each concept can be thought as the set of VLSs where it {\em holds}.

From the statements of the question, we have the hypotheses $lual\colon Laywer$ (Luana acts as lawyer), $leoal\colon Lawyer$ (Leonardo acts as lawyer) and $bal\colon Lawyer$ (Bruno acts as lawyer). Using the deductive system for $i\mathcal{ALC}$ first presented in \cite{HPR:2010b}, we can prove that Luana, Bruno and Leonardo can not act as lawyers. The proof below shows the Luana case, and similar deductions can be constructed for the others.

\begin{prooftree}
  \AxiomC{$lual\colon Police$}
  \AxiomC{$Police\subs\neg Lawyer$}
  \BinaryInfC{$lual\colon\neg Lawyer$}
  \AxiomC{$[lual\colon Lawyer]$}
  \BinaryInfC{$lual\colon\bot$}
  \UnaryInfC{$\neg(lual\colon Lawyer)$}
\end{prooftree}

\section{Conclusion and Future Works}

We presented a new data set with all Brazilian bar exams and their answer keys jointly with three Brazilian norms in LexML format. Furthermore, we also presented some preliminary experiments with the goal of constructing a system to pass in the OAB exam. We obtained reasonable results considering the simplicity of the methods employed.

For the next steps, many other experiment setups can be tested. For instance, we can construct the TF-IDF vectors using lemmas of the words, possibly increasing the similarities. We can also add edges between articles, considering that 10\% of our golden set includes more than one article as justification. 
We also plan to use the OpenWordnet-PT \cite{own-pt}, properly expanded with terms of the legal domain, following results from \cite{Sagri:2004tn}. 
In another direction, we need to increase the size of the golden set. Using crowd-sourcing websites to obtain more justifications from humans is a possibility. 

Many different proposals for data standards for the representation of laws are available. Why has a single standard not been largely adopted yet? We aim to explore the best candidate for encoding, possibly in collaboration with the LexML team, the remaining normative documents that we will need to cover all areas of the OAB exams. 

Finally, the results of the experiments presented here clearly show that we need ‘deep’ linguistic processing to capture the meaning of natural language utterances in representations suitable for performing inferences. That will require the use of a combination of linguistic and statistical processing methods, possibly using results from \cite{Quaresma:2005vw}. The final objective is to obtain formal representations, encoded in $i\mathcal{ALC}$ or another variant, from the texts ready for formal reasoning.

\bibliographystyle{plain}
\bibliography{oab}

\end{document}